# Stochastic Dynamic Power Dispatch with High Generalization and Few-Shot Adaption via Contextual Meta Graph Reinforcement Learning

Bairong Deng, Tao Yu, *Senior Member, IEEE*, Zhenning Pan*, *Member IEEE*, Xuehan Zhang*, Yufeng Wu, Qiaoyi Ding

*Abstract*—Reinforcement learning is an emerging approaches to facilitate multi-stage sequential decision-making problems. This paper studies a real-time multi-stage stochastic power dispatch considering multivariate uncertainties. Current researches suffer from low generalization and practicality, that is, the learned dispatch policy can only handle a specific dispatch scenario, its performance degrades significantly if actual samples and training samples are inconsistent. To fill these gaps, a novel contextual meta graph reinforcement learning (Meta-GRL) for a highly generalized multi-stage optimal dispatch policy is proposed. Specifically, a more general contextual Markov decision process (MDP) and scalable graph representation are introduced to achieve a more generalized multi-stage stochastic power dispatch modeling. An upper meta-learner is proposed to encode context for different dispatch scenarios and learn how to achieve dispatch task identification while the lower policy learner learns context-specified dispatch policy. After sufficient offline learning, this approach can rapidly adapt to unseen and undefined scenarios with only a few updations of the hypothesis judgments generated by the meta-learner. Numerical comparisons with state-of-the-art policies and traditional reinforcement learning verify the optimality, efficiency, adaptability, and scalability of the proposed Meta-GRL.

## Nomenclature

### Abbreviations

| | |
|---|---|
| CMDP | Contextual Markov decision process |
| DED | Dynamic economic dispatch |
| DRL | Deep reinforcement learning |
| ES | Energy storages |
| Meta-GRL | Energy storages |
| Meta-RL | Meta reinforcement learning |
| MDP | Markov decision process |
| RL | Reinforcement learning |
| RE | Renewable energy sources |
| TG | Thermal generations |

### Indices, Sets, and Parameters

| | |
|---|---|
| $i,j$ | Indices for system buses |
| $ij$ | Indices for lines |
| $t$ | Scheduling stages |
| $G_{ij}/B_{ij}$ | Conductance/susceptance of line $ij$ |
| $N_{RE}/N_{TG}/N_{ES}$ | Set of REs/TGs/ESs |
| $N_{bus}/N_{load}/N_{line}$ | Set of buses/loads/lines |
| $\Omega_i$ | Set of buses connected to bus $i$. |
| $P_{max}^{TG}/P_{min}^{TG}$ | Maximum/minimum active power output of TGs |
| $Q_{max}^{TG}/Q_{min}^{TG}$ | Maximum/minimum reactive power output of TGs |
| $P_D^{TG}/P_U^{TG}$ | Down/up ramping rate limit of TGs |
| $P_{max}^{RE}/P_{min}^{RE}$ | Maximum/minimum forecasted power of REs |
| $P_d^{ES}/P_c^{ES}$ | Discharging/charging power of ES |
| $P_{d,max}^{ES}/P_{c,max}^{ES}$ | Maximum discharging/charging power of ES |
| $T$ | Set of scheduling stages. $t$=1, 2, …, $|T|$ |
| $U_{max}/U_{min}$ | Maximum/minimum voltage amplitude |
| $\theta_{max}/\theta_{min}$ | Maximum/minimum phase angle |
| $\beta_{TG,a}/\beta_{TG,b}/\beta_{TG,c}$ | Cost coefficient of TGs |
| $\beta_{RE}$ | Penalty coefficient of REs |
| $\beta_{ES}$ | Degradation cost coefficient of ES |
| $\eta_{ES,c}/\eta_{ES,d}$ | Charging/discharging efficiency of ES |

### Variables

| | |
|---|---|
| $C(t)$ | Cumulative costs over dispatch horizon |
| $C^{TG}(t)$ | Cost function of TGs |
| $C^{RE}(t)$ | Penalty term of abandoned REs |
| $C^{ES}(t)$ | Cost function of ESs |
| $E_{max/min}^{ES}(t)$ | Maximum/minimum energy stored in ES |
| $E^{ES}(t)$ | Energy stored in ESs |
| $P^D(t)$ | Active load demand |
| $Q^D(t)$ | Reactive load demand |
| $P^L(t)$ | Active line transmission power |
| $Q^L(t)$ | Reactive line transmission power |
| $U(t)$ | Voltage of buses |
| $\theta(t)$ | Phase angle of buses |
| $P^{TG}(t)$ | Active power output of TGs |
| $Q^{TG}(t)$ | Reactive power output of TGs |
| $P^{RE}(t)$ | Active power output of REs |
| $P^{ES}(t)$ | Active power output of ESs |
| $E^{ES}(t)$ | Energy stored in ESs |

This work was supported by the National Natural Science Foundation of China (52207105), Guangdong Basic and Applied Basic Research Foundation(2023A1515011598) and the Natural Science Foundation of China-Smart Grid Joint Fund of State Grid Corporation of China(U2066212)
Bairong Deng, Tao Yu, Zhenning Pan, Yufeng Wu, Qiaoyi Ding are with the College of Electric Power, South China University of Technology, Guangzhou 510640, China
(e-mail: 1401989233@qq.com, taoyu1@scut.edu.cn, scutpanzn@163.com, wuyuffeng@163.com, 2426428967@qq.com);
Xuehan Zhang is with College of Electrical Engineering and Automation, Fuzhou University, Fuzhou 350116, China (e-mail: xh_zhang@fzu.edu.cn.)

* Corresponding authors. Zhenning Pan and Xuehan Zhang contributed equally to this work.

## I. INTRODUCTION

I N recent years, the increasing penetration of renewable energy (RE) and flexible distributed energy sources, such as energy storage (ES), has resulted in substantial challenges in the operation of power systems [1]. Specifically, the high penetration of RE leads to significantly different characteristics from conventional generation, particularly in terms of significant intermittency and uncertainties [2]. ES and storage-like devices, e.g., electric vehicles, provide precious flexibility but introduce strong temporal coupling constraints. As a result, it is crucial for system operators to develop an intelligent and dynamic dispatch policy to maintain a supply-demand balance accounting for multivariate uncertainties. Herein, the term policy represents a rule or function of how to make decisions according to the system state.

### A. Reinforcement Learning in Multi-stage Stochastic Power Dispatch

Extensive studies have been carried out to handle the multi-variate uncertainties in power system dispatch. Many of them incorporate stochastic or robust techniques to unit commitment and economic dispatch models in day-ahead or two-stage perspective. However, in day-ahead and two-stage formulations, the time-dependent uncertainties are assumed to be observed in a single interval, instead of being gradually revealed over time in the intra-day stage [3]. This ignores how the system will adapt to the realized uncertainty sequentially and may result in a large deviation from the optimal dispatch. Therefore, with growing uncertainties, it is necessary to transform the dynamic power system dispatch into a multi-stage sequential optimization problem and develop its computationally efficient policy, which allows sequentially given decisions based on the observed system state without knowledge of future stochastic processes [4].

Computing the optimal multi-stage dispatch policy is quite challenging due to the extremely high and nonenumerable state and decision space. For the sake of compromise, the lookahead model, also known as model predictive control [5]-[7], is most widely adopted for intra-day dispatch. However, a preferable lookahead policy usually requires dynamically updated and accurate forecasts over a long horizon, which may be unavailable. Besides, the computational burden brought by a long horizon may also hinder its application in short-term decision-making. Alternatively, intelligent data-driven approaches, reinforcement learning (RL), or approximate dynamic programming in typical, can tackle the high-dimension multi-stage problem which may be intractable by traditional methods like backward dynamic programming. They have been exploited to enhance decision-making ability in dynamic dispatch recently. Some authors leverage differential dynamic programming and stochastic dual dynamic programming schemes to respectively solve multi-period optimal power flow and multistage economic dispatch problems [8]-[9]. Some reinforcement learning methods are utilized to achieve efficient distributed economic dispatch in microgrids [10]. However, the aforementioned focused on how to master a specific power dispatch task, generalization to the changed environment and new task/scenario has received comparatively less attention.

### B. Low Generalization of Existing Reinforcement Learning based Power Dispatch

Although RL approaches work well in theory and simulation, their practical success in power dispatch is still quite limited. The major weakness that hinders their applicability and feasibility is their *low generalization ability*. Specifically, most existing studies treat stochastic processes in the power system as a Markov Decision Process (MDP) and conduct end-to-end uncertainty management. The uncertainty is usually assumed to follow a given distribution or be a noise term. This brings the following two major drawbacks:

Firstly, the environment and state transition of real-world power systems may be distributed dispersedly, resulting in diverse dispatch policies. For example, network topology changes due to fault and maintenance, and seasonal and climate change lead to different RE generation patterns. Our previous work [11] indicated that the more stochastic the environment, the lower the optimality of the RL dispatch result. The dispersed distribution of uncertainties leads to various tasks which may not be suitable for single MDP modeling. Unluckily, most existing RL-based power dispatch can hardly be generalized cross-task, resulting in *i)* a meticulous classification of historic data should be handled additionally to specify tasks for learning. Due to high dimensional task space, this is unrealistic in practice. *ii)* a specific agent needs to be pre-trained for a specific task, and a large amount of training samples and time are required to achieve an acceptable policy.

Secondly, a performant RL dispatch scheme highly depends on the consistency between training and testing environments. Real-world power systems may encounter scenarios which are rare, out-of-sample, or cross-task. Therefore, if the agent has insufficient generalization, the dispatch decision quality may degrade greatly and even become infeasible when the deviation between training and testing environments occurs. A highly generalized scheme will adapt its multi-stage policy with incomplete online observations of uncertainties' realization. Unluckily, most of the existing research applying RL to power dispatch neglected such issues by simply assuming consistency [8]-[11].

### C. Current Attempts to Improve Generalization of Reinforcement Learning based Power Dispatch

Several studies have adopted some state-of-the-art techniques to improve the generalization of RL-based dispatch schemes. Our previous works [4] and [12]-[13] have incorporated risk measurement and transfer learning to improve the dispatch quality of RL in unseen scenarios. However, a risk-averse or robust RL can only be generalized to cases where a limited mismatch between training and testing environments occurs instead of cross-task. While transfer learning highly depends on the state similarity to update a dispatch policy, sometimes negative transfer could happen. Some studies used multi-task deep reinforcement learning to enhance the generalization of all task policies. However, the related works have to pre-define tasks or sacrifice the optimality under specific scenarios to generalize across tasks, such as encoding the topology as an additional state and exploiting the optimal knowledge from source tasks to accelerate the search rate in new tasks [14]-[15].

Recently, a meta-learning framework has emerged to enhance the generalization of RL. In general, meta-learning constructs a higher-order learning architecture over traditional end-to-end RL. The learned meta-knowledge is used to achieve fast adaption of dispatch policy in highly stochastic power systems. Due to these advantages, meta-learning has been adopted for other sequential optimization problems like robot control. Our previous attempt in [16] used meta-learning in load monitoring. Model-agnostic meta-learning (MAML) is used in [17] to train initial model parameters of RL-based home energy optimizer. However, MAML is more like a training or tuning technique instead of a learning model. If the task space is large, finding good initial parameters may be difficult. Contextual meta-learning is another meta-learning technique. It encodes each task and develops a task-based policy. Compared with MAML, it enjoys better interpretability and adaptability. Recently, contextual meta-learning has been applied to grid emergency control [18] and delayed voltage recovery [19]. Since the dispatch problem is non-stationary, i.e., state transition evolves over time, how to adapt dispatch policy in intra-day according to the incomplete observed state trajectory is much more difficult than in the cases in [18] and [19]. In fact, to the best of our knowledge, the study and application of Meta-RL are still in a very early stage, especially in the field of power dispatch. The aforementioned methods have not been utilized to solve multi-stage stochastic power dispatch.

### D. Contributions and Paper Organization

Aiming at improving the low generalization and practicality of current RL used in power dispatch, this paper proposes a novel contextual meta graph reinforcement learning (Meta-GRL) method for multi-stage optimal dispatch policies. The main novelties and contributions are summarized as follows:

(1) A more generalized contextual MDP (CMDP) is introduced to formulate multi-stage stochastic power dispatch. A graph representation of system states is employed. In contrast with traditional MDP, such formulation not only captures different load and RE patterns but also generalizes to topology changes.

(2) A Meta-GRL is proposed to solve a highly generalized policy. Under this hierarchical learning structure, the upper meta-learner learns to encode context for each sample (task identification), while the lower policy learner learns context-specified dispatch policy. In such a way, a universal dispatch policy that accommodates diverse and changing power system environments can be obtained.

(3) An adaptive discriminator is proposed and trained to fully utilize meta-knowledge. By transforming probabilistic context inference into deterministic feature matching, an online cross-task adaptation of dispatch policy under incomplete system trajectory observation of new scenarios can be achieved.

The remainder of this paper is organized as follows: Section II describes the problem and formulates the Contextual MDP. In Section III, the Meta-RL methodology is formally introduced.

In Section IV, case studies are presented. Section V presents a discussion. Finally, conclusions are made in Section VI.

## II. PROBLEM DESCRIPTION AND MODELLING

### A. Problem Description and Formulation

The multi-stage stochastic power dispatch is studied in this paper. The unit commitments are assumed to be made at the beginning of dispatch. In each stage, the system operator observes the newly revealed system state and uncertainties, including load profiles and RE generation, then develops dispatch decisions of the current stage by adjusting the output of different controllable resources, e.g., conventional generators, energy storages, and flexible loads. The objective is to find the optimal policy, which minimizes the measured accumulative cost over stages.

$$\textbf{Min } \rho(\sum_{t \in T} C(t)) = \rho(\sum_{t \in T}(C^{\text{TG}}(t) + C^{\text{RE}}(t) + C^{\text{ES}}(t)))$$

$$\begin{cases} C^{\text{TG}}(t) = \sum_{i \in N_{\text{TG}}} (\beta_{\text{TG,a}} P_i^{\text{TG}}(t)^2 \Delta t + \beta_{\text{TG,b}} P_i^{\text{TG}}(t)\Delta t + \beta_{\text{TG,c}}) \\ C^{\text{RE}}(t) = \sum_{i \in N_{\text{RE}}} \beta_{\text{RE}}(P_{i,\max}^{\text{RES}}(t) - P_i^{\text{RES}}(t))\Delta t \\ C^{\text{ES}}(t) = \sum_{i \in N_{\text{ES}}} \beta_{\text{ES}}(P_{c,i}^{\text{ES}}(t) + P_{d,i}^{\text{ES}}(t))\Delta t \end{cases} \quad (1)$$

In this paper, for $\forall i \in N_{\text{TG}} \cup N_{\text{RE}} \cup N_{\text{ES}}$, we consider the most common case where the measurement operator $\rho[]$ is expectation $E$. However, some risk or robust measurements can also be used, as shown in our previous work in [20].

The dispatch decision at each stage should satisfy the following constraints:

Operation constraints of thermal generators:

$$\begin{cases} P_{i,\text{D}}^{\text{TG}} \leq P_i^{\text{TG}}(t) - P_i^{\text{TG}}(t-1) \leq P_{i,\text{U}}^{\text{TG}}, \\ P_{i,\min}^{\text{TG}} \leq P_i^{\text{TG}}(t) \leq P_{i,\max}^{\text{TG}}, \\ Q_{i,\min}^{\text{TG}} \leq Q_i^{\text{TG}}(t) \leq Q_{i,\max}^{\text{TG}}. \end{cases} \quad (2)$$

Operation constraints of renewable energy:

$$0 \leq P_i^{\text{RE}}(t) \leq P_{i,\max}^{\text{RE}}(t). \quad (3)$$

Operation constraints of energy storage systems:

$$\begin{cases} 0 \leq P_{d,i}^{\text{ES}}(t) \leq P_{d,i,\max}^{\text{ES}}, \\ 0 \leq P_{c,i}^{\text{ES}}(t) \leq P_{c,i,\max}^{\text{ES}}, \\ E_{i,\min}^{\text{ES}}(t) \leq E_i^{\text{ES}}(t) \leq E_{i,\max}^{\text{ES}}(t), \\ E_i^{\text{ES}}(t+1) = E_i^{\text{ES}}(t) + (\eta_{\text{ES,c}} P_{c,i}^{\text{ES}}(t) - P_{d,i}^{\text{ES}}(t)/\eta_{\text{ES,d}})\Delta t, \\ P_{c,i}^{\text{ES}}(t)P_{d,i}^{\text{ES}}(t) = 0. \end{cases} \quad (4)$$

Networks constraints:

$$\begin{cases} U_{i,\min} < U_i(t) < U_{i,\max}, \\ \theta_{i,\min} < \theta_i(t) < \theta_{i,\max}, \\ \displaystyle\sum_{i \in N_{TG}} P_i^{\mathrm{TG}}(t) + \sum_{i \in N_{RE}} P_i^{\mathrm{RE}}(t) + \sum_{i \in N_{ES}} P_i^{\mathrm{ES}}(t) = \sum_{i \in N_D} P_i^{\mathrm{D}}(t) + \sum_{i \in N_L} \Delta P_i^{\mathrm{L}}(t), \\ P_i^{\mathrm{TG}}(t) + P_i^{\mathrm{RE}}(t) + P_i^{\mathrm{ES}}(t) = \\ \qquad P_i^{\mathrm{D}}(t) + U_i(t) \displaystyle\sum_{j \in \Omega_i} U_j(t)(G_{ij}\cos\theta_{ij}(t) + B_{ij}\sin\theta_{ij}(t)), \\ Q_i^{\mathrm{TG}} = Q_i^{\mathrm{D}}(t) + U_i(t) \displaystyle\sum_{j \in \Omega_i} U_j(t)(G_{ij}\sin\theta_{ij}(t) - B_{ij}\cos\theta_{ij}(t)). \end{cases}$$

(5)

It should be noted that the problem in (1)~(5) is a general multi-stage power dispatch model. The proposed method is model-free and is scalable to other sequential optimization problems with different decision variables and objectives, e.g., multi-stage unit commitment.

Different from most existing research where the uncertainty is assumed to follow a fixed and given distribution or be a noise term, this paper considers a much more complex and practical case where the load and RE generation pattern in each sample may be diverse, and networks and topology can also differ due to line faults and maintenance. This is quite common in the current smart grid with increasing deployment of RE and soft equipment. This paper aims at developing a highly generalized dispatch policy for such cases. High generalization means that the performance of dispatch policy can be maintained in changing and diverse environments, and is easier to adapt to unseen scenarios.

### B. Generalized Dynamic Power Dispatch Model based on Contextual MDP

MDP is the most used model in multi-stage dynamic power dispatch optimization. As mentioned above, such a model is hard to generalize to the problem studied in this paper. Therefore, a more generalized contextual MDP compared with traditional MDP is introduced. MDP consists of $(\mathcal{S}, \mathcal{A}, \mathcal{T}(s_{t+1}|s_t, a_t), \mathcal{R})$, where $\mathcal{S}, \mathcal{A}, \mathcal{T}$, and $\mathcal{R}$ are the state space, action space, state transition, and reward, respectively. The specific action $a_t$ develops from policy $\pi(s_t)$. In contrast, contextual MDP (CMDP) is a tuple $(\mathcal{C}, \mathcal{S}, \mathcal{A}^C, \mathcal{T}^C(s_{t+1}|s_t, a_t), \mathcal{R}^C)$, where $\mathcal{C}$ is the context space and $C \in \mathcal{C}$. The major difference is that contextual MDP assumes a function $\mathcal{M}(C)$ mapping a latent context to the action space, system dynamics, and reward, correspondingly. In other words, a specific context defines a specific MDP tuple. In the dynamic power dispatch problem, $C$ contains diverse load and RE patterns and distributions, and even topology changes. These endogenous and exogenous differences may lead to significantly different dispatch policies. This is a quite reasonable setting in power dispatch optimization, because the state transition, cost function, and feasible operation region of the power system are affected by many underlying issues which are unobservable or hard to gather. For example, different seasons and weather dominate RE generation patterns obviously, and different types of dates and social behaviors affect load profiles.

The key difficulty is how to formulate and encode context for multi-stage dynamic power dispatch. Close contexts lead to similar system costs, dynamics, and corresponding dispatch policies. Previous approaches have used the system state similarity to divide different dispatch tasks. They used relatively straightforward and arbitrary ways. For example, our previous approaches [21] used netload similarity to measure system state similarity. However, this ignores the load and RE spatial distribution, and topology information is also not considered. System states with close net load may result in distinct policies. In this regard, the context for multi-stage power dispatch is defined as $C = [c_1, c_2, \dots c_T]^{\mathrm{T}}$ where $c_t = (s_t, a_t, r_t, s_{t+1})$. Such formulation not only considers the non-stationary and temporally dependent nature of power system state transition but also tries to integrate as much state information as possible for reasonable context definition.

A different context $C$ defines a different MDP, resulting in a different dispatch policy. If the dispatch policy performs well for $\forall C \in \mathcal{C}$, the policy is highly generalized. That is, a generalized dispatch policy equals a context-specified dispatch policy under contextual MDP formulation. However, finding such dispatch is quite challenging, because the context classification is not known in advance and the agent has to identify context from raw historic samples and learn a context-specified policy simultaneously. This issue will be tackled by the novel Meta-GRL method proposed in the following section.

## III. METHODOLOGY

### A. Overall Framework of Meta Graph Reinforcement Learning

This paper introduces the Meta-GRL algorithm for the multi-stage dynamic dispatch problem modeled by CMDP. We aim at finding a highly generalized dispatch policy which adapts to various and diverse scenarios of power systems. If the context-specified dispatch policy can dynamically adapt over the context space $C$ and demonstrate the optimal performance across various training tasks, the dispatch policy can be generalized to previously unseen tasks. The context-specified dispatch policy is defined as $\pi_\theta(s_t, C): \mathcal{S} \times \mathcal{C} \longrightarrow \mathcal{A}$, where $c$ can be extracted to embeddings $z$ ($C$ can be characterized by $z$), in which policy can be reformulated as $\pi_\theta(a|s, C(z))$. It is clear that the dispatch decision offered by $\pi_\theta$ not only depends on the revealed system state but also on the context in which the system is.

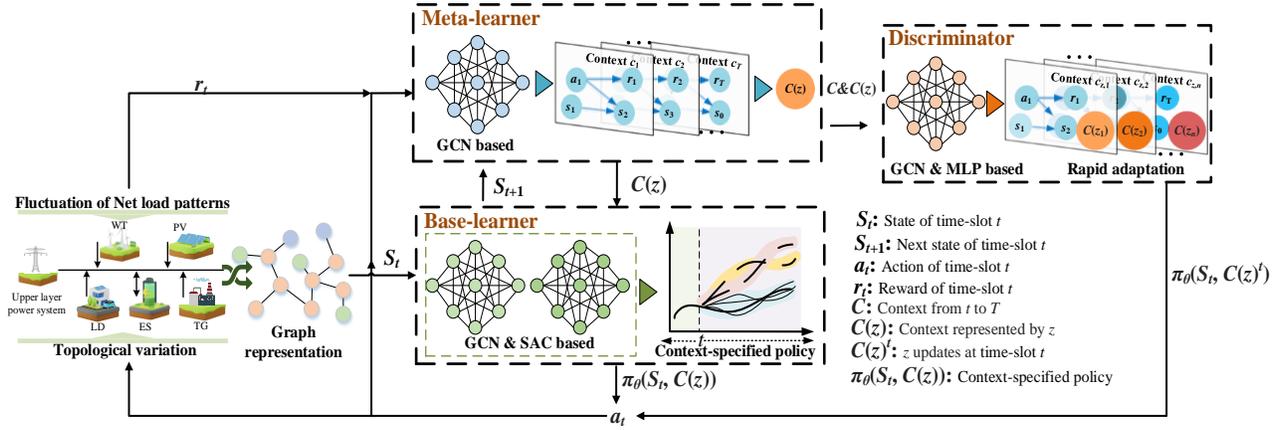

Fig. 1. Overall Framework of Meta Graph Reinforcement Learning.

As shown in Fig. 1, compared with traditional RL that solves policy via MDP, Meta-GRL that solves CMDP for dynamic power dispatch contains two distinct and hierarchical parts, i.e., the meta-learner and base-learner. Specifically, the meta-learner tries to distinguish and encode each context while the base-learner undertakes a context-specified dispatch policy learning. Fig. 1 plots the learning scheme of Meta-GRL. Its overall goal is to find the optimal strategy that maximizes the objective function of the stochastic dynamic power dispatch problem and the dispatch performance considering various and diverse scenarios over context space, as

$$\pi^* = \arg \max_{\pi} E_{(S_t, a_t, C(z)) \sim \rho_{\pi}} \{ \sum_t r(S_t, a_t) \} \quad (6)$$

In the meta-training phase, the meta-learner and base-learner work cooperatively to optimize the universal dispatch policy $\pi_{\theta}(S_t, C(z))$ and to maximize the rewards of all training samples. Specifically, the meta-learner assigns hypothesis judgments of context to various samples. These assignments remain constant throughout an iteration step of meta-training. Subsequently, given the embedded context, the base-learner learns the optimal context-specified policy that maximizes corresponding cumulative rewards. After receiving the feedback cumulative reward, the meta-learner and base-learner simultaneously optimize the parameters of latent context space $\mu$ (meta-learner) and the policy parameters $\theta$ (base-learner). In such a way, the meta-learner accurately discerns the discrepancy of context and the base-learner effectively tackles various tasks based on the latent probabilistic embeddings of context. While in the meta-testing phase, the model of the base-learner is frozen, the meta-learner further seeks to find a better context identification for improving the base-learner's performance under testing samples.

Context identification by the meta-learner has a certain degree of similarity with dispatch scenarios' classification in existing research. However, the key difference is that the objective of the meta-learner is to find the best context identification to help the base-learner form a universal policy which generalizes to diverse dispatch tasks.

When Meta-GRL is properly trained and ready for online application, the meta-learner has to observe an entire trajectory of system state, i.e., $[(S_1, a_1, r_1, S_2), \dots (S_t, a_t, r_t, S_{t+1}), \dots (S_T, a_T, r_T, S_0)]$, and infers the latent context where the power system is. Then the base-learner can choose a more suitable policy to tackle dynamic dispatch by embedding context. To further enhance the online adaptability of Meta-GRL, an adaptive discriminator is introduced to infer the latent context sequentially with an incomplete observation of $S_t$ trajectory. That is, at time-slot $t$, Meta-GRL is capable of adjusting its policy dispatch only according to the observation of $[(S_1, a_1, r_1, S_2), \ldots (S_t, a_t, r_t, S_{t+1})]$. In other words, Meta-GRL rapidly adjusts policy outcomes during an interaction round, relying on a restricted set of trajectory data. This adjustment influences the trajectory path, leading to a transition to a new subsequent state, $S'_{t+1}$. Consequently, this process enhances the grid's security and stability by implementing an improved initialization strategy.

### B. Graph Representation of the Power System State

As shown in Fig. 1, the meta-learner, base-learner, and adaptive discriminator all require a proper presentation of the power system state. Most existing research used a vector or matrix-based state representation, which relies on Euclidean space measurements. Nonetheless, the spatial distribution of load and RE, as well as topology information, may not be well captured. Inspired by our previous successful attempts in [22], a graph representation of the power system state is employed. Its advantages lie in the inherent consistency with the power system structure and the easily incorporated multi-variate data. A graph data is expressed as $G(Adj, Eig)$ in Fig. 2. The adjacency matrix ($Adj$) illustrates the connection relationships among buses in the given topology. Meanwhile, the eigenmatrix ($Eig$) consolidates the state features of each bus at time-slot $t$, aligning with the composition of the state space.

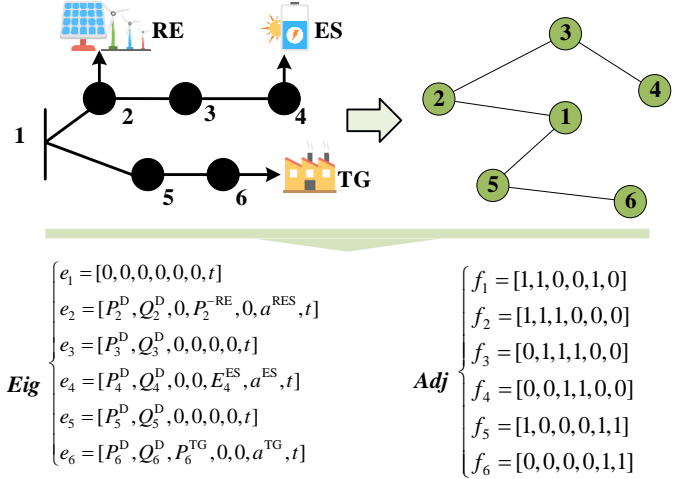

$$Eig \quad \begin{aligned} e_1 &= [0,0,0,0,0,0,t] \\ e_2 &= [P_2^D, Q_2^D, 0, P_2^{RE}, 0, a^{RES}, t] \\ e_3 &= [P_3^D, Q_3^D, 0,0,0,0,t] \\ e_4 &= [P_4^D, Q_4^D, 0,0, E_4^{ES}, a^{ES}, t] \\ e_5 &= [P_5^D, Q_5^D, 0,0,0,0,t] \\ e_6 &= [P_6^D, Q_6^D, P_6^{TG}, 0,0, a^{TG}, t] \end{aligned} \qquad Adj \quad \begin{aligned} f_1 &= [1,1,0,0,1,0] \\ f_2 &= [1,1,1,0,0,0] \\ f_3 &= [0,1,1,1,0,0] \\ f_4 &= [0,0,1,1,0,0] \\ f_5 &= [1,0,0,0,1,1] \\ f_6 &= [0,0,0,0,1,1] \end{aligned}$$

Fig. 2. The use of a graph in conveying the knowledge of the spatial distribution of net load pattern and topology information.

Illustrated in Fig. 2 is an example using the graph representation of the distribution network. In this context, the feature dimension of $Eig(n,m)$, denoted as $m$, is expandable and effectively captures the comprehensive feature information of each node. Simultaneously, the $Adj(n,n)$ matrix mirrors the connection relationships among $N$ nodes, facilitating real-time updates to capture topology variations at any given moment. Furthermore, this paper leverages a graph neural network with feature extraction capabilities to achieve information aggregation for $Eig$ and $Adj$.

### C. Base-learner design

After the graph representation of the system state is constructed, a graph convolution network (GCN) is employed to gather information and generate a new undirected graph $G(Adj, Eig')$, which serves as input to the meta-learner. The process of extracting features from graph data can be viewed as an aggregation of information from neighboring nodes, and this can be expressed as (7):

$$x_i^{(k)} = \sigma \left( \sum_{j \in N(i)} \frac{1}{\sqrt{\deg(i)} \cdot \sqrt{\deg(j)}} \cdot \Theta \cdot e_{i,j} x_j^{(k-1)} \right) \quad (7)$$

where $x_i^{(k)}$ is the feature of the graph node $i$ after $k$-th convolution layer, $\deg(i)/\deg(j)$ is the degree of node $i/j$, representing the total number of edges associated with this node. $\Theta$ denotes the weight matrix, and $e_{i,j}$ is the weight of an edge of the graph, respectively.

Therefore, a base-learner based on Soft Actor Critic (SAC) [23] makes corresponding network modifications by adding graph convolution to the network layers, which can be adapted to the input of graph features. Further, the key idea behind SAC is to maximize the expectation of cumulative rewards and also the entropy of the policy distribution. By maximizing entropy, SAC promotes optimal, diverse, and exploratory policies to ensure thorough exploration of all possible actions under different embeddings $z$ and to prevent getting trapped in local optima. Therefore, the optimal formulation and modified Bellman equations of policy $\pi_\theta(a|s,z)$ are shown in (8)

$$\pi_\theta(S_t, C(z)) = \arg\max_\pi E_{(S_t, a_t, C(z)) \sim \rho_\pi} \left\{ \sum_t r(S_t, a_t) \right.$$
$$\left. + \gamma E\{V_{t+1}(S_{t+1}, C(z)) | (S_t, C(z)), a_t + \alpha H[\pi[\cdot|(S_t, C(z))]]\} \right\} \quad (8)$$

$$Q(s_t, z), a_t) = r(s_t, a_t) + \gamma E_{S_{t+1}, z, a_{t+1}} \{Q[(s_{t+1}, z), a_{t+1})$$
$$- \alpha \ln[\pi[a_{t+1} | (s_{t+1}, z)]]\} \quad (8)$$

where $\rho_\pi$ is the distribution of state-action trajectory put forward by $\pi$. $H$ is the entropy term of action in the state $(s_t, C(z))$ and $\alpha$ is an intensity coefficient indicating the proportion of exploration ability. $Q[(s_t, C(z)), a_t]$, fitted through the nonlinear mapping function of the neural network, is a value of taking action $a_t$ in the state $(s_t, C(z))$, which replaces the effect on the expectation of the future. $V_t(s_t, C(z))$ is the mean value of rewards for reaching state $(s_t, C(z))$ at time-slot $t$.

The base-learner comprises two sets of networks: the Critic and the Actor. The Critic's parameters are updated by minimizing the residual $J_Q(\vartheta)$, as depicted in equation (9)

$$J_Q(\vartheta) = E_{(s_t, a_t, s_{t+1}, C(z)) \sim D, a_{t+1} \sim \pi_\theta} \left\{ \frac{1}{2} \begin{bmatrix} V_\vartheta(s_t, a_t, C(z)) - r(s_t, a_t) \\ -\gamma V_{\bar\vartheta}(s_{t+1}, a_{t+1}, C(z)) \\ +\alpha\gamma \ln(\pi_\theta(a_{t+1}|s_{t+1}, C(z))) \end{bmatrix} \right\} \quad (9)$$

where $\vartheta$ represents the Critic network parameter, $\theta$ is the Actor network's parameter, and $\bar\vartheta$ denotes the target Critic network parameter. The target Critic involves delayed updates of Critic parameters aimed at stabilizing the learning process. The target network parameter undergoes updates every 1000 interaction rounds following a specific method, with $\lambda$ representing the coefficient update rate, as depicted in equation (10)

$$\vartheta \leftarrow \lambda \vartheta + (1-\lambda)\overline{\vartheta} \tag{10}$$

The residuals of the Actor network are obtained by using the minimized KL scatter and multiplying by the coefficient $\alpha$, which are shown in equation (11):

$$J_\pi(\theta) = E_{s_t \sim D, a_t \sim \pi_\theta}[\alpha \ln(\pi_\theta(a_t|s_t, C(z)) - Q_\vartheta(s_t, a_t, C(z))] \tag{11}$$

### D. Meta-learner design

To facilitate adaptation, Meta-GRL must effectively reason about context distributions related to MDPs in dynamic power dispatch. Consequently, the meta-learner is configured to represent context by embedding $z$, extracting trajectory information from the given context. When encountering rare or unseen scenarios, the meta-learner is tasked with making hypothetical judgments and updating its belief regarding context distributions through trial and error. This aligns with the concept of posterior sampling, which maintains a posterior distribution over possible MDPs by acting optimally. Therefore, inspired by Probabilistic Embeddings for Actor Critic (PEARL) [24], this paper uses a Graph Convolutional Networks (GCN) network parameterized by $\omega$ to incorporate context inference into meta-learner and to further learn the posterior $p(z|c)$ for $z$. In a generative approach, this is achieved by maximizing returns through the context-specified policy. Assuming the objective is to maximize log-likelihood, we formulate the resulting variational lower bound as

$$E_\zeta\left[E_{z \sim q_\omega\left(z|c^\zeta\right)}\left[\beta D_{\text{KL}}\left(q_\omega\left(z|c^\zeta\right)\|\,p(z)\right)\right]\right] \tag{12}$$

where $p(z)$ is the distribution of Gaussian prior probability over $z$. The KL divergence term, constraints of mutual information between $z$ and $c$, can be elucidated to reduce the asymmetry of the difference between prior $p(z)$ and posterior $p(z|c)$ representing $q_\omega(z|c)$, which forces $z$ to contain only information from the $c$.

This design enables the meta-learner to be expressive enough to capture the minimal sufficient statistics of context-relevant information without modeling irrelevant dependencies. Thus, our method directly infers a posterior over the latent context $C(z)$, encoding the MDP to optimize its reconstruction, and optimal context-specified policies are subsequently optimized accordingly.

### E. Meta Training

In summary, the main components of Meta-GRL are the meta-learner and the base-learner. As shown in Fig.3, Our meta-training procedure, utilizes training scenarios to acquire a prior over $C(z)$. It captures the distribution of context and efficiently leverages experience to infer new tasks. After a predefined number of interaction rounds, the meta-learner initially extracts $c_t$ from the experience buffer in batch form and transforms $c_t$ into a graph representation that combines topological variations and net load pattern data. Subsequently, the meta-learner assesses $C(z)$ by encoding a hypothetical judgment of the scenario through the prior. $C(z)$ is then assumed to be a component of the state that is not normally observable, attached to the next interaction round. This process is equivalent to adding context-specified features to the base-learner. Building upon this foundation, the base-learner optimizes its strategy to maximize cumulative rewards across rounds. The meta-learner, in turn, uses this information to optimize its network parameters and update its judgments of trajectories.

Furthermore, in the design of the training process, Meta-GRL adopts a strategy of segregating the data used to train the meta-learner from that used to train the base-learner. This separation aims to enhance sampling efficiency and guarantee accurate inference of contexts from minimal interactions. A replay buffer $B$ is structured to accommodate experiences of all sampled scenarios for the training of the base-learner. Non-sequential data is sampled from this replay buffer $B$ to calculate the loss and update the parameters. To maintain the on-policy property of the meta-learner, a replay buffer $M$ with limited capacity is structured to store and encode $C$ in sequential form. The constraint on the volume of buffer $M$ is intended to expedite the exploration of adaptive latent probabilistic embeddings and prevent redundant attempts.

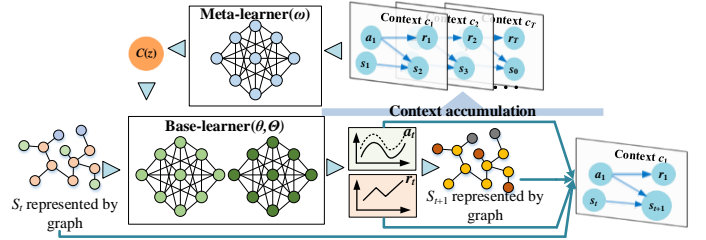

Fig. 3. The process of meta-training.

### F. Meta Testing

During the meta-testing phase, the procedures of Meta-GRL can be classified into three progressive parts. Firstly, the policy-learner executes actions on new tasks while setting $C(z)$ to the initialized value. It indicates the absence of potential task-specific characteristics. Secondly, the meta-learner samples $z$ from the prior distribution, enabling the base-learner to explore in a similarly structured and temporally extended manner. The third step involves utilizing the previously mentioned collected system state trajectory to update the posterior distribution. This process allows for a coherent and sustained exploration. It persists until a predetermined number of test rounds are completed to meet real-time dispatch response requirements. Consequently, when confronted with an unfamiliar context, such as unseen load and RE generation patterns, the meta-learner refines context hypotheses through a limited number of interactions. This refinement aims to achieve an optimal initialization policy for scheduling tasks in an unprioritized scenario.

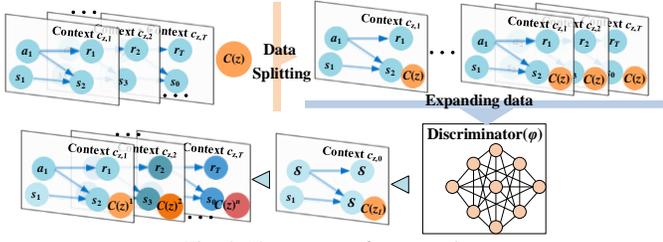

Fig. 4. The process of meta-testing.

### G. Discriminator Learning for Online Dispatch Policy Adaptation

An adaptive discriminator is introduced to Meta-GRL to achieve fast online adaption of dispatch policy to stochasticity. Different from the meta-learner which can only adjust dispatch policy after observing a complete horizon from $t=1$ to $T$, instead, the adaptive discriminator allows Meta-GRL to update context inference dynamically according to the realized system state trajectory.

The function of the adaptive discriminator is to predict the most likely context which the current system is going through. The discriminator undergoes modification by incorporating a GCN atop a multilayer perceptron (MLP), enabling its adaptation to the graph representation of context. Following both meta-training and meta-testing, we gather information pairs $[C, C(z)]$ after model convergence. Subsequently, we systematically decompose $C$ into finite-observation pairs in a time-ordered incremental manner. This process establishs a many-to-one mapping with $C(z)$. The objective of this step is to employ the principles of supervised learning, enabling the discriminator to fit incomplete mappings between $C$ and $C(z)$. This process aids the base-learner in obtaining updated $C(z)^t$ at different scheduling time slots within a round. This, in turn, accelerates the generation process of the initialization strategy. Consequently, the policy learner produces actions through the policy $\pi_\theta(a|s, C(z)^t)$, swiftly adjusting the real-time scheduling of power. This transformation converts probabilistic online inference into deterministic embedding matching, allowing for immediate adaptation to unseen scenarios.

### H. Overall Procedure of Meta-GRL

As given in Algorithms 1 & 2, there are two important parts, offline learning and online adaptation: the meta-learner and policy-learner learn about the embeddings and optimal policy during meta-training(update $\omega$, $\theta$, $\Theta$), and the discriminator enhances the identification (update $\varphi$).

---

**Algorithm 1**: Offline meta-training of meta-learner and policy-learner

1: **Step 1: Initialization**
2: Initialize the replay buffer B and b and learning rates $\alpha_1$, $\alpha_2$, $\alpha_3$. Generate a set of training interaction experiences of multi-stage sequential optimization task $p(\zeta)$. Input $\omega$, $\theta$, $\Theta$, $\vartheta$.
3: **Step 2: Meta training**
4:    **while** not done(when $t = T$, done is True) **do**
5:      Initialize context **c**= { }
6:      **for** $m =1,…, M$ **do**, $M$ is total sampling interaction round
7:        Sample $\mathbf{z} \sim p(\mathbf{z})$, $q_\omega(\mathbf{z}|\mathbf{c})$
8:        Gather data from $\pi_\theta(a|s, \mathbf{z})$ and add to B & b

---

9:        Update $\mathbf{c} = \{(s_j, a_j, s_{j+1}, r_j)\}_{j:1…T} \sim$ B & b (only reserve latest context)
10:      **End for**
11:    **for** step in training steps **do**
12:      Sample context $\mathbf{c} \sim$ B and context $\mathbf{c'} \sim$ b
13:      Graph representation $\mathbf{g_c}=g_\vartheta(\mathbf{c})$
13:      Sample $\mathbf{z} \sim q_\omega(\mathbf{z}| \mathbf{g_c})$, $q_\omega(\mathbf{z}| \mathbf{g_c'})$
14:      Calculate $\acute{L}_{actor}(\mathbf{g_c}, \mathbf{z})$, $\acute{L}_{critic}(\mathbf{g_c}, \mathbf{z})$
15:      Calculate $\acute{L}_{KL}(\mathbf{g_c'}, \mathbf{z}) = \beta D_{KL}(q(\mathbf{z}| \mathbf{g_c'})||r(\mathbf{z}))$
16:      Update $\vartheta$, $\omega$ by $\alpha_1 \nabla_{\vartheta\omega}(\acute{L}_{critic}(\mathbf{g_c}, \mathbf{z}) + \acute{L}_{KL}(\mathbf{g_c'}, \mathbf{z}))$
17:      Update $\vartheta$ by $\alpha_2 \nabla_\vartheta \acute{L}_{critic}(\mathbf{g_c}, \mathbf{z})$
18:      Update $\theta$ by $\alpha_3 \nabla_\theta \acute{L}_{actor}(\mathbf{g_c}, \mathbf{z})$
19:    **End for**
20: **End while**

---

**Algorithm 2**: Online meta-testing of meta-learner and discriminator

1: **Step 1: Discriminator learning**
2: Generate cumulative context $c_{\hat{e}}$ produced by well-trained meta-learner in chronological increments as training sample D, initialize learning rates $\alpha_4$
3:    **for** step in training steps **do**
4:      Sample batch of D
5:      Calculate $\acute{L}_{dr}(D)$
6:      Update $\varphi$ by $\alpha_4 \nabla_\varphi \acute{L}_{dr}$ (D)
7:    **End for**
8: **Step 2: Meta test**
9:    Initialize context $\mathbf{c_{test}}$= { }
10: **for** $i$ in testing steps **do**
11:    Graph representation $g_{c_i}=g_\vartheta(c_i)$
11:    Obtain $z \sim q_\varphi(\mathbf{z}| g_{c_i})$
12:    Roll out policy $\pi_\theta(a|s, \mathbf{z})$ to collect data $\mathbf{c_{test}} = \{(s_j, a_j, s_{j+1}, r_j)\}_{j:1…T}$
13:    Accumulate context $\mathbf{g_c} = \mathbf{g_c} \cup \mathbf{g_{c,test}}$
14:    Update $\varphi$ by $\alpha_4 \nabla_\varphi \acute{L}_{dr}$ ($\mathbf{g_c}$, $\mathbf{z}$)
14: **End for**
**Output**: the well-trained discriminator and policy for unseen scenarios.

---

## IV. CASE STUDIES

### A. Settings of Case 1

Case study 1 is conducted on intra-day dispatch of a modified 39-bus network including TGs, REs, and an ES. The dispatch interval is set to 15 minutes, and thus a whole day involves 96 stages. Detailed parameters of the case can be found in [25]. The number of segments for flexible resources is provided in Appendix B. Both training and testing samples in simulations can be found at https://github.com/JWdung/Operation-data-of-actual-power-grid. The parts highlighted in red in Fig. 5 represent potential topology changes. Possible topological changes, line outages, load switching, and unit commitment are assumed to occur before the time slot 0. The initial power outputs of flexible resources are set to 0 and the initial state of charge of ES is set to 50%.

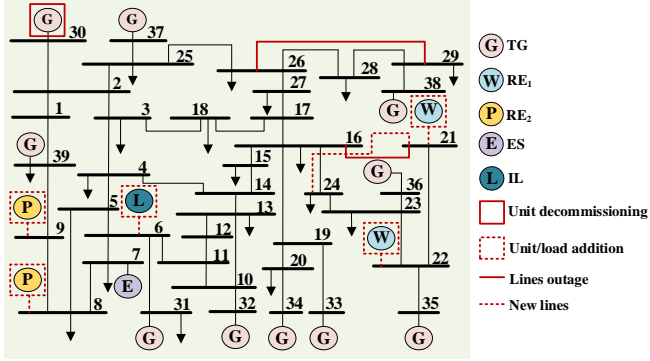

Fig. 5. Modified 39-bus networks with flexible resources.

Since it is hard to obtain the exact optimal policy, this paper uses the optimal posterior solution (OPS) [25] of each sample to compare the optimality of Meta-GRL and other algorithms. OPS can be obtained by directly solving the deterministic problem after all uncertainties are revealed [13]. The measurement of optimality is calculated as follows:

$$\text{Optimality} = \frac{F_{\text{OPS}}}{F} \times 100\% \tag{13}$$

where $F_{\text{OPS}}$ is the objective value generated by OPS and $F$ is the objective value generated by Meta-GRL and other algorithms.

Meta-GRL is modeled by the deep learning library Torch and Rlkit. All the simulations are implemented on Intel (R) Core (TM) i7-10700F CPU @ 2.90 GHz 2.90 GHz processor, 16.0 GB of on-board RAM, and a graphics card NVIDIA GeForce GT 730.

### B. Performance of Meta-GRL in Case 1

In this subsection, five scenarios composed of distinct load and RE generation patterns are manually generated to verify whether Meta-GRL can perceive the differences among scenarios and develop a context-specified policy, and whether Meta-GRL finds a policy which generalizes to unseen scenarios. In addition to the variations in RE and load patterns, S1-S5 also incorporate random topological changes. For example, there is a probability that the line between buses 16 & 21, 26 & 29 will be interrupted, as shown in Fig. 5. Fig. 6 shows the error bars and the expected net load curves of different patterns. S1-4 is utilized as training scenarios for meta-GRL to verify the interpretability of the task inference function of the meta learner. And S5 is utilized as a testing/online adaptation scenario to verify the generalization results of the model.

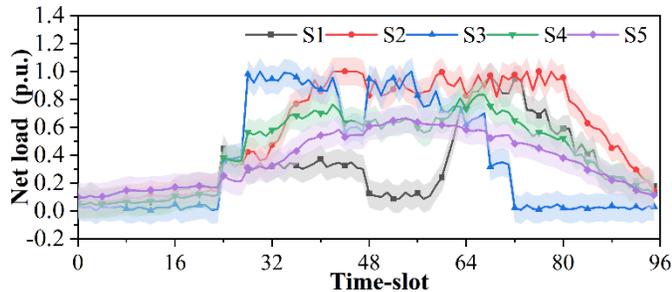

Fig. 6. Characteristic curves of net load

The approaches used for comparison include OPS by solving the deterministic problem of each sample, traditional DRL under the SAC framework trained by different scenarios, and the look-ahead dispatch, i.e., model predictive control (MPC). Specifically, SAC (S1-S4) is SAC trained by mixed samples from S1 to S4, and SAC (S1) is SAC trained by samples from S1. The goal of MPC is to minimize total costs considering future uncertainties over the prediction horizon. In this paper, the MPC with different prediction horizons is denoted as MPC-n. For example, MPC-24 indicates an MPC policy that incorporates prediction information for the next 24 stages. The forecast is assumed to be the expectation value of uncertainties and the problem is solved by the interior point optimizer.

First, to verify whether the meta-learner can capture the latent context of samples from different scenarios and help the base-learner to make dispatch decisions more elegantly, the performances of Meta-GRL and traditional DRL (SAC) are compared. Meta-GRL interacts with a training environment that contains mixed samples from S1-S4. The results are shown in both Table. I and Fig. 9.

Table. I lists the results of the different algorithms under different scenarios. It can be seen that traditional DRL is sample-dependent. They only work well if the testing environment is consistent with the training environment, and the performance cannot be ensured when facing out-of-distribution samples. Although they can be trained by mixed samples from different scenarios (SAC (S1-S4)), only sub-optimal policy can be achieved and the results are poor in all testing scenarios. This is because traditional RL methodologies based on MDP cannot reflect the latent factors which may determine the MDP parameters. They are not applicable to the power dispatch if the system operation state varies greatly. In contrast, Meta-GRL outperforms SAC (S1-S4) under all scenarios. It achieves over 90% optimality in all scenarios, i.e., S1: 93.04%, S2: 93.55%, S3: 91.95%, and S4: 93.28%. The reason is twofold: first, Meta-GRL adopts a more generalized CMDP modeling that is universal to different MDP models. Second, the meta-learner of Meta-GRL can distinguish samples well from different scenarios by extracting latent contexts among them. By embedding $C(z)$, the base-learner can learn a context-specified dispatch policy which can be generalized and adapted to the power system with complex uncertainties.

Fig. 9 plots the convergence curves of different algorithms under different scenarios. It is evident that Meta-GRL outperforms SAC (S1/S2/S3/S4) in terms of convergence rate and optimality in segmented scenarios. The reason behind this lies in the limitations of traditional DRL, which relies solely on a restricted set of observation states. When unable to discern the potential trajectory trend, it requires more optimization time to discover the optimal solution. However, the latent contexts $C(z)$ extracted by the meta-learner of Meta-GRL are used as new state features, which help the base-learner to analyze trajectory information, find more optimal solutions more quickly, and therefore accelerate the convergence. The further reason for this lies in the effective design of meta-learning. That is, the mutual information constraint makes $C(z)$ characterize only the trajectory-related information, which more clearly portrays the context.

For a more intuitive and visual understanding of why Meta-

GRL outperforms traditional RL under complex uncertainties, the samples of different patterns are visualized in a two-dimensional form using the dimensionality reduction method called t-distributed stochastic neighbor embedding (t-SNE), as shown in Fig 8. It is clear that according to t-SNE, the samples are clustered into 4 scenarios which are consistent with Fig. 7. The latent context probabilistic embeddings $C(z)$ extracted by the meta-learner are also visualized using the t-SNE technique, as shown in Fig. 8. It is clear that the context identification exhibits almost the same results as Fig. 7, i.e., the samples from the same scenario are also classified into similar contexts. Thus, when encountering different scenarios, base-learners can make targeted power dispatch by fully considering the differences of their latent MDPs.

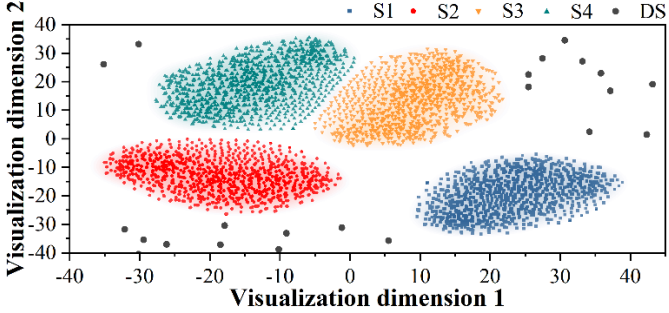

Fig. 7. Distribution of S1-4 in two-dimensional visualization space

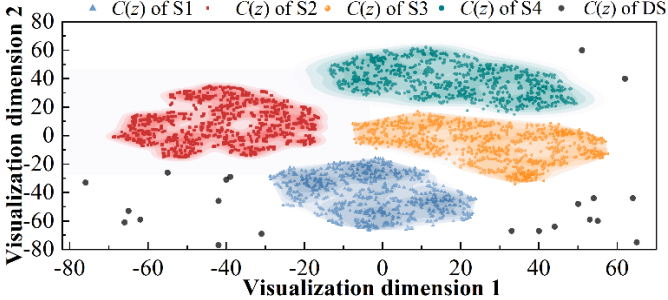

Fig. 8. Distribution of $C(z)$ in two-dimensional visualization space

To further validate the rapid adaptive capability of Meta-GRL to non-prior tasks, we presented randomly generated samples to assess the generalization characteristics of Meta-GRL in more extreme or risky discrete scenarios (DS). The two-dimensional spatial distribution of these unordered samples does not fall into any of the previously defined S1-5 , as illustrated in Figs. 7 and 8. Upon comparing the hidden task feature results generated by the meta-learner, it becomes evident that Meta-GRL can effectively recognize and adapt to unordered discrete scenes. This highlights the model's ability to handle diverse and unfamiliar scenarios, showcasing its robust adaptability beyond predefined tasks.

Eventually, to verify the generalization of Meta-GRL to unseen scenarios, a new scenario with different load and RE characteristics from S1-S4 is established and named S5. As shown in Fig. 10, traditional DRL trained by S1-S4 cannot adapt to S5. It has to be retrained to obtain a reliable dispatch policy, which requires a substantial amount of training time and computing resources. Such a disadvantage limits the practicality of traditional DRL when applied to power dispatch of actual power systems. In contrast, there is a significant reduction in the number of iterations required for Meta-GRL to achieve convergence on S5. Even after a small number of interactions, Meta-GRL provides superior results compared to fully trained SAC and guarantees 88.45% optimality compared to OPS. This highlights the high generalization of Meta-GRL to unseen scenarios. Further, the dispatch results of different algorithms are also compared in Fig. 11, it can be observed that the results obtained by Meta-GRL are much closer to OPS compared with the results of traditional DRL (SAC). Specifically, as shown in Fig 11 (a) & (c), it can be observed that Meta-GRL exhibits superior sub-optimal performance for TGs and ES, whose variation tendencies and amplitude range closer to OPS. Considering the ample capacity of REs in the system and the goal of maximizing their utilization, both OPS and Meta-GRL aim to align the RE outputs with the real-time power upper bound, as depicted in Fig. 11. (c).

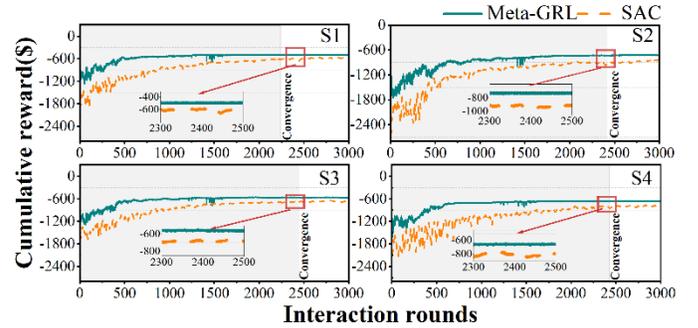

Fig. 9. Cumulative reward variations of S1-4 in offline training

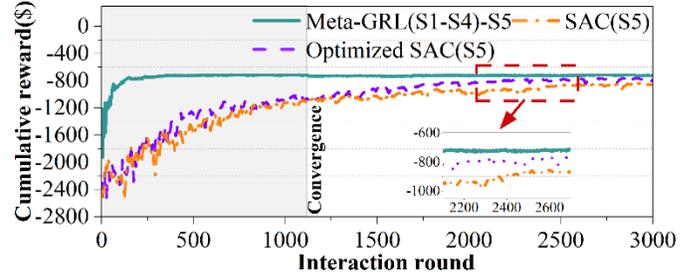

Fig. 10. Cumulative reward variations of S5 in an online adaptation

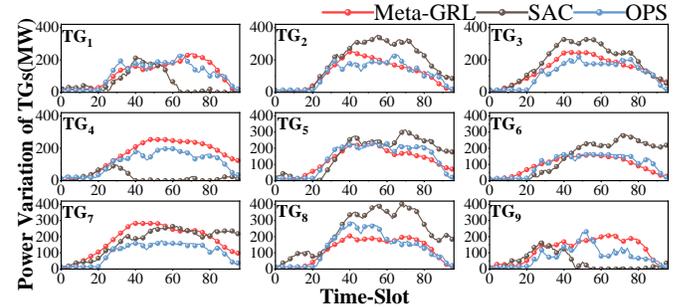

(a) Power variation of TGs

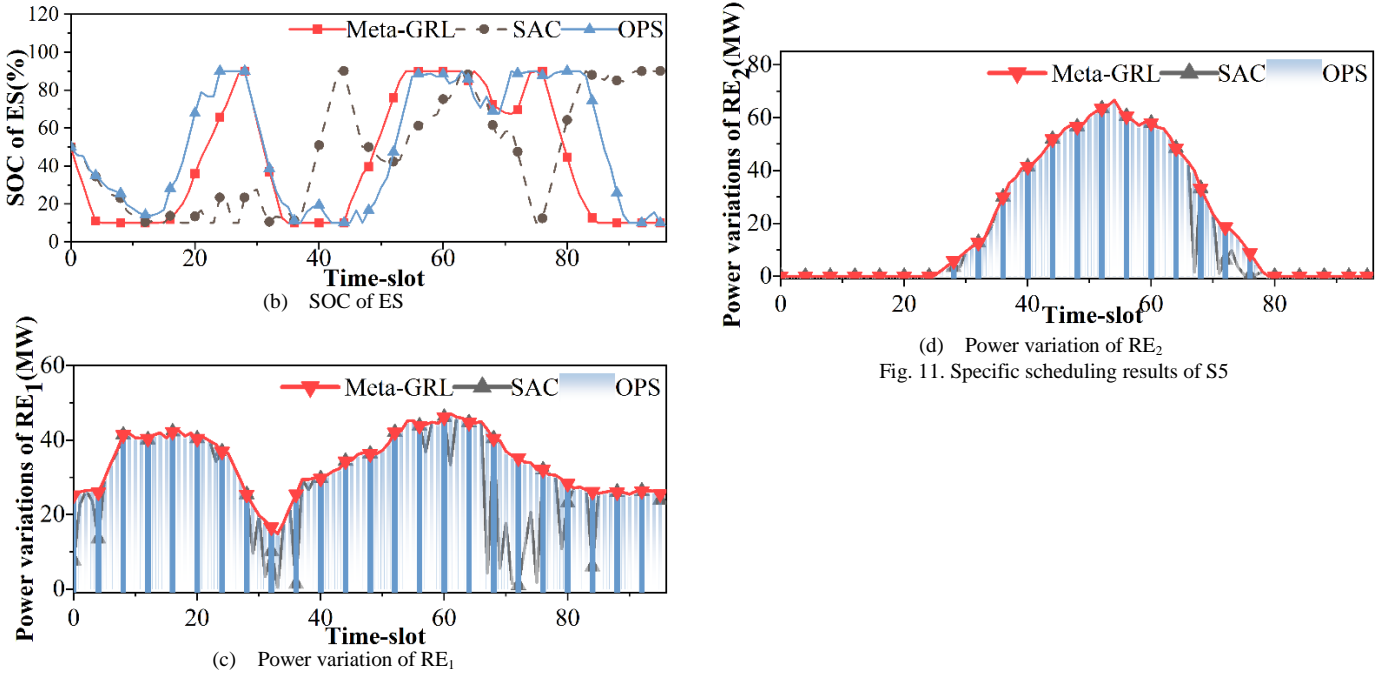

(b) SOC of ES

(d) Power variation of RE$_2$
Fig. 11. Specific scheduling results of S5

(c) Power variation of RE$_1$

TABLE I
PERFORMANCE OF DIFFERENT ALGORITHMS

| | OPS | Meta-GRL | SAC(S1-S4) | SAC(S1) | SAC(S2) | SAC(S3) | SAC(S4) | SAC(S5) | MPC-48 | MPC-24 | MPC-12 | MPC-4 |
|---|---|---|---|---|---|---|---|---|---|---|---|---|
| Objective of S1/day($) | **4.534*10⁶** | **4.873*10⁶** | 8.576*10⁶ | **5.803*10⁶** | 7.303*10⁶ | 8.146*10⁶ | 7.247*10⁶ | 7.161*10⁶ | 4.845*10⁶ | 4.847*10⁶ | 4.856*10⁶ | 4.900*10⁶ |
| Objective of S2/day($) | **6.716*10⁶** | **7.179*10⁶** | 9.286*10⁶ | 9.771*10⁶ | **8.473*10⁶** | 1.238*10⁷ | 1.059*10⁷ | 9.942*10⁶ | 6.755*10⁶ | 6.784*10⁶ | 6.803*10⁶ | 6.806*10⁶ |
| Objective of S3/day($) | **5.072*10⁶** | **5.516*10⁶** | 8.925*10⁶ | 6.744*10⁶ | 7.752*10⁶ | **6.208*10⁶** | 7.226*10⁶ | 7.523*10⁶ | 5.345*10⁶ | 5.354*10⁶ | 5.365*10⁶ | 5.760*10⁶ |
| Objective of S4/day($) | **6.054*10⁶** | **6.490*10⁶** | 9.124*10⁶ | 9.183*10⁶ | 9.573*10⁶ | 8.162*10⁶ | **7.623*10⁶** | 8.516*10⁶ | 6.325*10⁶ | 6.414*10⁶ | 6.552*10⁶ | 6.938*10⁶ |
| Objective of S5/day($) | **6.364*10⁶** | **7.195*10⁶** | 8.210*10⁶ | 8.566*10⁶ | 8.179*10⁶ | 8.632*10⁶ | 9.501*10⁶ | **8.061*10⁶** | 6.553*10⁶ | 6.643*10⁶ | 6.701*10⁶ | 6.894*10⁶ |

## C. Case 2

In this subsection, we adopt a more practical and realistic case to further verify the effectiveness of Meta-GRL in addressing dynamic power dispatch under high uncertainties. This case derives from the 2022 Artificial Intelligence Application Contest of power system – Track 1: Real time Collaborative Dispatching Based on Reinforcement Learning [26], which was held by State Grid Corporation of China. The tackled dynamic power dispatch contains 288 dispatch stages and the time interval is 5 minutes. The training data consists of 100,000 power flow data over one year, which originates from historical operation data of an actual power grid. Besides, the environment consists of contingencies, e.g., huge fluctuations in load and RE generation, and line faults. The metrics for evaluating the performances of algorithms include penalty of line overruns and balancing power unit, positive incentive of RE unit consumption, operating costs of TGs, etc. This assessment spans various aspects, including grid security, low carbon impact, and economic considerations.

Fluctuations in net load, unit start/stop, the probability of faults occurring, and other power grid environment settings are restricted to the simulators sealed by the competition organizers. To assess the generalizability of the proposed algorithm, we established thresholds for the soft overload rate and heavy overload rate, enabling the differentiation of various scenarios. The mixed multiple scenarios were then presented to both Meta GRL and SAC agents for interactive learning, with the objective of training robust and adaptable agents to face unexpected power grid situations. Fig. 12 illustrates the variation curves of cumulative rewards during an interaction round (due to the undisclosed network topology data, OPS is unable to obtain). It is evident that incomplete decisions during the initial adaptation period result in the balance unit exceeding its limits when faced with sudden start/stop events of generation units or abrupt changes in net load patterns. This, in turn, can lead to system collapse and suspension of operations, impacting cumulative rewards. However, through further exploration, Meta-GRL successfully identified a viable solution for this scenario, achieving a full operating rate within a single interaction round.

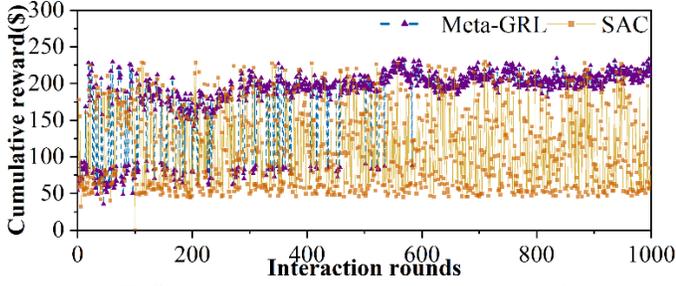

Fig. 12. Cumulative reward variations of different algorithm

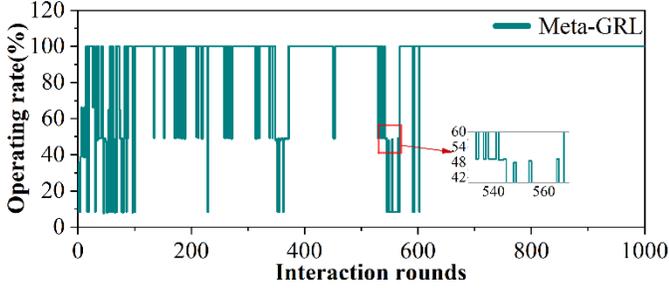

Fig. 13. Operating rate of Meta-GRL

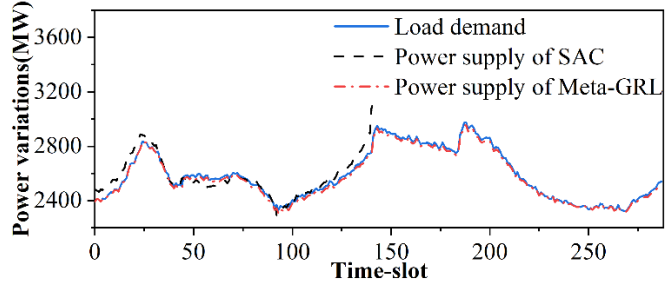

Fig. 14. Power variations of total load and main power

Finally, this paper compares the load profile within the 1000 th round, as described above, with the generation schedules of the two methods. It is evident that DRL (SAC) disengages from operation when the total load undergoes drastic changes, and the disparity between its decision result and the total load exceeds the balance constraint. In contrast, Meta-GRL adeptly tracks load changes and promptly adjusts its strategy to fulfill the power balance requirements.

## V. DISCUSSION

In the context of high renewable energy penetration and the presence of flexible resources, ensuring the feasibility of strategies in complex operating scenarios becomes crucial for maintaining power system reliability and economy. This paper introduces Meta-GRL as a potential solution to address the limitations of CRL in handling these challenges. Meta-GRL offers practicality by leveraging latent embeddings extracted from trajectory-level data to identify scenario differences, directly influencing downstream decision-making. Furthermore, Meta-GRL demonstrates strong generalization and adaptability by providing high-quality initial strategies for unseen scenarios, effectively adapting to various uncertainties. Additionally, Meta-GRL exhibits high efficiency by leveraging offline learning to generate high-quality online solutions, eliminating the need for extensive forecast data, abundant computational resources, and retraining time.

While Meta-GRL demonstrates promising generalization in uncertain environments, there is room for enhancing its optimality in initial policies for unseen scenarios. Improving the optimality of Meta-GRL presents an interesting avenue for future research. Additionally, the current formulation of topology uncertainties occurring before the initial scheduling time-slot 0 may not capture more realistic issues, such as topological variations happening at specific points within the scheduling cycle, including TG shutdown/start constraints or operating failures. Validating whether the proposed Meta-GRL method can maintain generalization and provide feasible solutions in such cases is an important consideration for future investigations. Expanding the application of Meta-GRL to more complex unit commitment problems would further enhance its practicality. Moreover, exploring whether the meta-learner based on trajectory-level contexts can provide different latent embeddings to guide decision-making adjustments during moments of concentrated data uncertainties, such as peak load demand, presents an intriguing direction for further exploration.

## CONCLUSIONS

This paper focuses on tackling the challenges of low generalization and practicality of existing reinforcement learning based power dispatch, a novel contextual meta-GRL for multi-stage optimal dispatch policy is proposed. The main findings are summarized as follows:

1) The Meta-GRL framework, employing the CMDP scheme and graph representation, effectively integrating multidimensional information in the form of encoded context. In contrast with traditional MDP, Meta-GRL formulation not only captures different load and RE patterns, but also generalizes to topology changes.

2) Under this hierarchical learning structure of the Meta-GRL framework, the upper meta-learner learns to encode context for each sample to achieve task identification, while the lower policy learner learns context-specified dispatch policy. This paper showcases the robust adaptability of Meta-GRL beyond predefined tasks.

3) The context-specified policy generated by Meta-GRL consistently outperforms the DRL's across all domains in terms of both asymptotic performance and sample efficiency. This validates that the contexual meta learning framework has enhanced the generalization of RL when dealing with power dispatch.

4) Meta-GRL enables rapid trajectory-level adaptation online, requiring only a few interaction rounds to adapt to unseen scenarios. This process is further accelerated by the setting of the discriminator. This approach effectively conserves computing resources and time, while still providing more optimal initialization strategies.

Overall, the proposed Meta-GRL framework showcases its ability to effectively address the challenges with high-dimensional uncertainty in power dispatch, achieving improved generalization, convergence, sample efficiency, and real-time adaptability while conserving computational resources and time.

REFERENCES

[1]  Z. Pan, T. Yu, J. Li, et al. "Multi-Agent Learning-Based Nearly Non-Iterative Stochastic Dynamic Transactive Energy Control of Networked Microgrids," *IEEE Transactions on Smart Grid.*, vol. 13, no. 1, pp. 688-701, Jan. 2022.

[2]  Q. Hou, E. Du, N. Zhang and C. Kang, "Impact of High Renewable Penetration on the Power System Operation Mode: A Data-Driven Approach, " *IEEE Transactions on Power Systems.*, vol. 35, no. 1, pp. 731-741, Jan. 2020.

[3]  J. D. Lara, O. Dowson, K. Doubleday, B. -M. Hodge and D. S. Callaway, "A Multi-Stage Stochastic Risk Assessment With Markovian Representation of Renewable Power," *IEEE Transactions on Sustainable Energy.*, vol. 13, no. 1, pp. 414-426, Jan. 2022.

[4]  Z. Pan, T. Yu, J. Li, K. Qu, B. Yang, "Risk-averse real-time dispatch of integrated electricity and heat system using a modified approximate dynamic programming approach", *Energy.*, Volume 198, 2020,117347,ISSN 0360-5442.

[5]  W. Gu, Z. Wang, Z. Wu, Z. Luo, Y. Tang and J. Wang, "An online optimal dispatch schedule for CCHP microgrids based on model predictive control", *2017 IEEE Power & Energy Society General Meeting.*, Chicago, IL, USA, 2017, pp. 1-1.

[6]  H. Qiu, W. Gu, Y. Xu, Z. Wu, S. Zhou and G. Pan, "Robustly Multi-Microgrid Scheduling: Stakeholder-Parallelizing Distributed Optimization", IEEE *Transactions on Sustainable Energy.*, vol. 11, no. 2, pp. 988-1001, April 2020.

[7]  H. Zou, Y. Wang, S. Mao, F. Zhang, and X. Chen, "Distributed online energy management in interconnected microgrids", *IEEE Internet of Things Journal.*, vol. 7, no. 4, pp. 2738–2750, Apr. 2020.

[8]  A. Agarwal and L. Pileggi, "Large Scale Multi-Period Optimal Power Flow With Energy Storage Systems Using Differential Dynamic Programming", *IEEE Transactions on Power Systems.*, vol. 37, no. 3, pp. 1750-1759, May 2022.

[9]  Y. Lan, Q. Zhai, X. Liu and X. Guan, "Fast Stochastic Dual Dynamic Programming for Economic Dispatch in Distribution Systems", *IEEE Transactions on Power Systems.*, vol. 38, no. 4, pp. 3828-3840, July 2023.

[10]  W. Liu, P. Zhuang, H. Liang, J. Peng and Z. Huang, "Distributed Economic Dispatch in Microgrids Based on Cooperative Reinforcement Learning", *IEEE Transactions on Neural Networks and Learning Systems.*, vol. 29, no. 6, pp. 2192-2203, June 2018.

[11]  B. Deng, J. Chen, Q. Ding, et al. "Multi-task Deep Reinforcement Learning Optimal Dispatchng Based on Grid Operation Scenario Clustering", *Power System Technology.*, vol. 47, no. 3, pp. 978-990, Mar 2023.

[12]  X. Zhang, Z. Xu, T. Yu, B. Yang and H. Wang, "Optimal Mileage Based AGC Dispatch of a GenCo", *IEEE Transactions on Power Systems.*, vol. 35, no. 4, pp. 2516-2526, July 2020.

[13]  X. Zhang, C. Li, B. Xu, Z. Pan and T. Yu, "Dropout Deep Neural Network Assisted Transfer Learning for Bi-Objective Pareto AGC Dispatch", *IEEE Transactions on Power Systems.*, vol. 38, no. 2, pp. 1432-1444, March 2023.

[14]  Y. Pei, J. Zhao, Y. Yao and F. Ding, "Multi-Task Reinforcement Learning for Distribution System Voltage Control With Topology Changes", *IEEE Transactions on Smart Grid.*, vol. 14, no. 3, pp. 2481-2484, May 2023.

[15]  J. Xie et al., "Imitation and Transfer Q-Learning-Based Parameter Identification for Composite Load Modeling", *IEEE Transactions on Smart Grid.*, vol. 12, no. 2, pp. 1674-1684, March 2021.

[16]  L. Xiong et al., "Meta-Reinforcement Learning-Based Transferable Scheduling Strategy for Energy Management", *IEEE Transactions on Circuits and Systems I: Regular Papers.*, vol. 70, no. 4, pp. 1685-1695, April 2023.

[17]  Y. He, F. Luo and G. Ranzi, "Transferrable Model-Agnostic Meta-learning for Short-Term Household Load Forecasting With Limited Training Data", *IEEE Transactions on Power Systems.*, vol. 37, no. 4, pp. 3177-3180, July 2022.

[18]  R. Huang et al., "Learning and Fast Adaptation for Grid Emergency Control via Deep Meta Reinforcement Learning", *IEEE Transactions on Power Systems.*, vol. 37, no. 6, pp. 4168-4178, Nov. 2022.

[19]  Y. Du et al., "Physics-Informed Evolutionary Strategy Based Control for Mitigating Delayed Voltage Recovery", *IEEE Transactions on Power Systems.*, vol. 37, no. 5, pp. 3516-3527, Sept. 2022.

[20]  Z. Pan, T. Yu, W. Huang, Y. Wu, J. Chen, K. Zhu, J. Lu, "Real-time dispatch of integrated electricity and thermal system incorporating storages via a stochastic dynamic programming with imitation learning", *International Journal of Electrical Power & Energy Systems.*,Volume 153,2023,109286,ISSN 0142-0615.

[21]  X. Zhang, T. Bao, T. Yu, B. Yang, C. Han, "Deep transfer Q-learning with virtual leader-follower for supply-demand Stackelberg game of smart grid", *Energy.*, Volume 133,2017,Pages 348-365,ISSN 0360-5442.

[22]  J. Chen, T. Yu, Z. Pan, M. Zhang, B. Deng, "A scalable graph reinforcement learning algorithm based stochastic dynamic dispatch of power system under high penetration of renewable energy", *International Journal of Electrical Power & Energy Systems.*, Volume 152,2023,109212,ISSN 0142-0615.

[23]  S. Wang, R. Diao, C. Xu, D. Shi and Z. Wang, "On Multi-Event Co-Calibration of Dynamic Model Parameters Using Soft Actor-Critic", *IEEE Transactions on Power Systems.*, vol. 36, no. 1, pp. 521-524, Jan. 2021.

[24]  K. Rakelly, A. Zhou, D. Quillen, C. Finn, S. Levine. "Efficient Off-Policy Meta-Reinforcement Learning via Probabilistic Context Variables", *International Conference on Machine Learning.*, vol 97, Jun.2019.

[25]  R. H. Byrne, T. A. Nguyen, D. A. Copp, B. R. Chalamala and I. Gyuk, "Energy Management and Optimization Methods for Grid Energy Storage Systems", *IEEE Access.*, vol. 6, pp. 13231-13260, 2018.

[26]  2022 Artificial Intelligence Application Contest of power system -- Track 1: Real time Collaborative Dispatching Based on Reinforcement Learning. https://aistudio.baidu.com/competition/detail/423